\documentclass[twocolumn, 11pt]{article}

\usepackage{microtype}
\usepackage{graphicx}
\usepackage{subfigure}
\usepackage{booktabs} 
\usepackage{siunitx}

\usepackage{hyperref}

\usepackage{arxiv}

\usepackage[utf8]{inputenc} 
\usepackage[T1]{fontenc}    
\usepackage{hyperref}       
\usepackage{url}            
\usepackage{booktabs}       
\usepackage{amsfonts}       
\usepackage{nicefrac}       
\usepackage{microtype}      
\usepackage{lipsum}		
\usepackage{graphicx}
\usepackage{natbib}
\usepackage{doi}
\usepackage{authblk}

\usepackage{amsmath}
\usepackage{amssymb}
\usepackage{mathtools}
\usepackage{amsthm}

\renewcommand{\vec}[1]{\mathbf{#1}}
\newcommand\zu{\vec{z}_\mathrm{u}}
\newcommand\zr{\vec{z}_\mathrm{r}}
\newcommand\xr{\vec{x}_\mathrm{r}}
\newcommand\xu{\vec{x}_\mathrm{u}}
\newcommand\xc{\vec{x}_\mathrm{c}}

\title{Measuring Unintended Memorisation of Unique Private Features in Neural Networks}

\date{} 					

\author[1, 3]{\small John Hartley}
\author[1, 2, 4]{\small Sotirios A. Tsaftaris}

\affil[1]{\footnotesize School of Engineering, University of Edinburgh, West Mains Rd, Edinburgh EH9 3FB, UK}
\affil[2]{\footnotesize The Alan Turing Institute, London, UK}
\affil[3]{\textit {john.hartley@ed.ac.uk}}
\affil[4]{\textit {s.tsaftaris@ed.ac.uk}}

\renewcommand{\shorttitle}{Measuring Unintended Memorisation of Unique Private Features in Neural Networks}

\hypersetup{
pdftitle={Measuring Unintended Memorisation of Unique Private Features in Neural Networks},
pdfauthor={John Hartley, Sotirios A. Tsaftaris},
}

\begin{document}

\twocolumn[
    \begin{@twocolumnfalse}
    \maketitle
\end{@twocolumnfalse}]
\begin{abstract}Neural networks pose a privacy risk to training data due to their propensity to memorise and leak information. Focusing on image classification, we show that neural networks also unintentionally memorise unique features even when they occur only once in training data. An example of a unique feature is a person's name that is accidentally present on a training image. Assuming access to the inputs and outputs of a trained model, the domain of the training data, and knowledge of unique features, we develop a score estimating the model’s sensitivity to a unique feature by comparing the KL divergences of the model’s output distributions given modified out-of-distribution images. Our results suggest that unique features are memorised by multi-layer perceptrons and convolutional neural networks trained on benchmark datasets, such as MNIST, Fashion-MNIST and CIFAR-10. We find that strategies to prevent overfitting (e.g.\ early stopping, regularisation, batch normalisation) do not prevent memorisation of unique features. These results imply that neural networks pose a privacy risk to rarely occurring private information. These risks can be more pronounced in healthcare applications if patient information is present in the training data.
\end{abstract}

\section{Introduction}
Deep Neural Networks (DNNs) are a powerful tool for classifying images \citep{LeCun1989, Krizhevsky2012-alexnet, He2016-resnet, Huang2017-densenet}. It is now commonly known that DNNs memorise training labels when the number of trainable parameters in the network is greater than the cardinality of the training set \citep{Zhang2021-understanding-mem}. This is the case for real, noisy, and random data, whether their labels have been shuffled or not \citep{Arplt2017-a-closer-look-at-memorization}.

\begin{figure}[ht]
    \centering
    \includegraphics{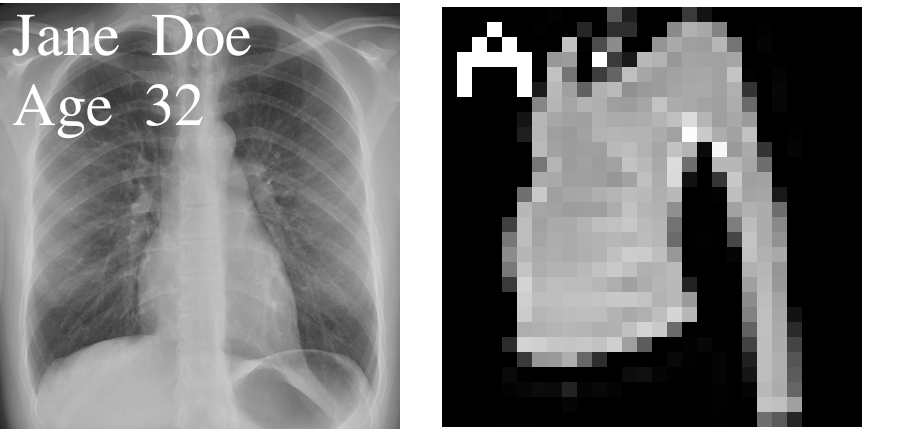}
    \caption{What will happen if a training dataset of X-ray images mistakenly contains a \textit{single}  image that shows the name of a patient? Will this unique information be exploited (in other words memorised) by a neural network? As we describe in the text, such a mistake is possible as data sanitisation is not always perfect. Here we explore these questions by injecting fake visual features in canary images from known benchmark datasets (an example canary image is shown on the right). We devise a memorisation score that assesses whether such \textit{unintended memorisation} is possible. We explore whether architecture, regularisation, and the canary image matter, respectively. The short answer is \textit{Yes}. [To alleviate any privacy concerns, this X-ray image is in the public domain \cite{wiki:Chest_radiograph}, and the patient name we use is fictional and commonly used in the media to denote an unknown individual.]}
    \label{fig:the-gist}
\end{figure}

Recent work by \citet{feldman-shorttaleaboutalongtail, feldman-what-neural-networks} has established theoretically and empirically that DNNs can achieve close to optimal generalisation error in image classification tasks data when examples, predominantly rare and atypical from long-tailed data distributions, are memorised. It has also been shown empirically for several benchmark datasets, that memorised examples exert a large influence on the network predictions for atypical but visually similar examples in the test set \citep{feldman-what-neural-networks}.

The memorisation of examples poses a privacy risk since information relating to the example is encoded directly in the weights of a neural network \cite{Golatkar2020}. For example, an adversary could construct a readout  function acting on the weights or network outputs to discover information about a given example \cite{shokri2017membership}. Data leakage is particularly problematic when datasets contain private information for which disclosure is limited. For example, DNNs used in healthcare may encode information about patients in their weights \citep{7163871, Zech2018-pneumonia-classification}, for which disclosure is limited in the EU by the General Data Protection Regulation (GDPR).


Herein, we focus on the \textit{unintended memorisation} of specific image features as opposed to examples or training labels. Although it has been shown that training examples are memorised \citep{Zhang2021-understanding-mem} it is not clear whether an example is memorised in its entirety or whether specific features of the image are memorised. This distinction is important since private features in an image could be memorised, and leaked. For example, hospitals frequently employ sanitisation processes to remove patient names when they appear overlaid on X-ray films (see example in Figure \ref{fig:the-gist}). But not all processes are foolproof, and occasionally an image with a patient's name will make it to the training dataset. A classifier trained on this data may misdiagnose other patients with the same name if those names also have not been removed. Another more likely risk is that this unintended presence may lead to incomplete extraction of the correct discriminative features from the image. Such a risk is similar to decision making based on spurious correlations, except that only a single spurious feature is contained in the dataset \citep{Zech2018-pneumonia-classification, Geirhos2020-shortcut-learning, Idrissi2021}.

We propose that unique feature memorisation can be detected through data leakage. Methods that exploit data leakage to uncover information about training data are called privacy attacks. One example is the membership inference attack (MIA) which finds whether an example is in the training set. This is achieved by exploiting a model's overconfidence on examples it has seen \citep{shokri2017membership, Sablayrolles2018, salem2018-ml-leaks, liu2020-have-you-forgotten, choquettechoo2021-label-only-mia}. Here instead we focus on the memorisation of unique features and not whole data. Adopting the terminology from \citet{jegorova2021survey} we will call this \textit{feature leakage}.

Feature leakage occurs in large language models (LLM) due to unintended memorisation \citep{carlini2019secret, Carlini2020}. Unique sequences, such as credit card numbers or rarely occurring phrases, are likely to be memorised by a model since they are random and cannot be \textit{learnt} from similar patterns in the rest of the training data.

We extend the notion of feature leakage in large language models to the supervised image classification setting. We consider a training set that contains a single private unique feature $\zu$ which is present in a single image $\vec{x}$, an example of which we show in Figure \ref{fig:the-gist}. We are interested in detecting whether $\zu$ is memorised by a neural network.

We aim to develop a black-box methodology to measure the memorisation of known unique private features in a trained model, assuming access only to the inputs and outputs of the model and to the unique private feature. We assume that we do not have access to the training data, but we do know its domain \footnote{For example, the domain of an X-ray dataset would be X-rays.}.


Our \textbf{main contributions} are summarised as follows:
\begin{itemize}
    \item A new setting of assessing unique feature memorisation in images.
    \item A score to assess the memorisation of unique features in neural networks where only access to the inputs/outputs and the unique feature is allowed.
    \item We find that neural networks memorise unique features injected into several benchmark datasets MNIST, Fashion-MNIST and CIFAR-10. This holds even when a single unique feature is present once in the dataset, and before overfitting in an average sense occurs.
    
    \item We find that the risk of such memorisation is not eliminated by adding explicit/implicit regularisation. On the contrary, memorisation is more likely.

    \item We hypothesise that hard-to-learn data influence the memorisation of features. We find that the degree of memorisation is weakly correlated with the self-influence score (a measure of learning difficulty) of the training image on which the unique feature is present.
\end{itemize}
\section{Related Works}

To the best our of knowledge, no previous study has investigated methods that detect memorisation of unique features in image classification models. However, existing research recognises the critical role that memorisation plays in the overfitting and generalisation of neural networks and the security implications of attacks using adversarial training examples. Below we outline work on memorisation that has provided encouraging findings to help us shape this work. We hypothesise that hard-to-learn examples can make memorisation of features more likely so we cover also related work on finding high influence examples.

\subsection{Memorisation}
The main objective of this work is to understand whether memorisation occurs for unique features that occur once per dataset and are located on a single training example. A secondary aim is to show that on average, the onset of this behaviour occurs before overfitting happens.

Previous research has established that over-trained, over-parameterised neural networks memorise training labels \citep{Zhang2021-understanding-mem}, with several metrics for assessing the memorisation of training labels having been previously introduced \citep{feldman-what-neural-networks, Jiang2021}.

It was shown that DNNs first learn common patterns in training examples, after which they memorise labels \citep{Arplt2017-a-closer-look-at-memorization, Kim2018a-gan-mem}. More recently, it has been shown that learning and memorisation occur simultaneously \citep{Liu2021}. These works also show that regularisation does not eliminate memorisation. We take inspiration from these works to explore whether the effects of regularisation are similar in unique feature memorisation.


Recent work by \citet{carlini2019secret} has established that large language models memorise unique phrases early in training. They show a black-box inference method for measuring memorisation in trained models. This work has inspired us to draw upon the use of canaries to measure the memorisation of unique features in images. We define a \textit{canary} as a training example that contains an artificially injected unique feature. We follow their idea to also investigate memorisation before \textit{overlearning} occurs.

Property inference attacks 
attempt to learn some group property/feature of the dataset. For example, what proportion of people in the training set wear glasses? \citep{ateniese2013-property-inference, ganju2018-property-inference}. These attacks are typically whitebox and proceed by inference on model weights. Feature memorisation, as we investigate here, can be viewed as an extreme property inference attack where a unique feature, a person who wears glasses, occurs only once in the dataset. Such approaches, however, cannot address unique feature memorisation since labelling the training weights requires knowledge of whether the feature was memorised or not.

We also draw upon the idea of concept activation vectors (CAV), as introduced by \citet{kim2018interpretability-cav}, to infer the sensitivity of the network to the feature we are interested in vs.\ random features that we are not. However, instead of image concepts such as stripes, we focus on unique features and use only the outputs of the trained model as opposed to the internal activations of the network. Our memorisation score also does not require the label of the unique feature.

\subsection{Finding high influence examples}

Memorisation can occur when DNNs generalise from examples that are mislabelled or belong to sub-populations/long-tails within classes to samples in the test set \citep{feldman-shorttaleaboutalongtail, feldman-what-neural-networks}. Similarly, research into long-tailed learning establishes that models have a worst-group accuracy as a result of class imbalances or spurious correlations in datasets \citep{Liu2020a, zhang2021deep, Liu2021b}.

We hypothesise that unique features injected on such examples are more likely to be memorised. These examples can be found using influence functions, or proxy functions to influence functions \citep{Koh2017, pmlr-v80-katharopoulos18a, carlini2019prototypical, Ghorbani2019, Toneva2019, feldman-what-neural-networks, Garima2020, Guo2020-fastif, Baldock2021, Harutyunyan2021, Jiang2021}. Influence functions are generally expensive to compute for deep learning models since they rely on the difference in the expected model predictions of models which are trained with and without the example under test. However, approximations using sampling, model approximations or alternatives based on learning dynamics are more feasible. We make extensive use of the computationally efficient TracIn score to estimate the influence of an example on its prediction by a trained model \citep{Garima2020}.

\section{Methodology}

\subsection{Problem statement}

DNNs are known to memorise training examples in image classification tasks. We propose that DNNs also independently memorise unique features of image examples even when they occur extremely rarely in the training data. This behaviour is a privacy concern when those features are private, such as a name or an address. To answer this question we discuss a metric that approximates the memorisation of a unique feature that occurs once in a training set.

We discuss a score that can verify the privacy of a model trained on data that may or may not contain private information. Our setting assumes the possibility of a fallible preprocessor that while tasked to remove private features fails on a single case.

Concretely, we limit our discussion to the following case: we measure whether a given unique feature is memorised given black-box access to the model (inputs and outputs), and without access to training examples or to the training labels.

We define a unique feature as a unique set of pixels in a single image within the training set. This feature could be a name, a postcode, or other information which should not be disclosed. We do not know a priori whether the feature is in the dataset or not since we cannot catch errors in the preprocessor, and therefore we cannot construct a dataset without the unique feature.\footnote{Thus, ruling out leave-one-out measures for memorisation.} However, we do assume we know the form that the unique feature takes since we are aware of the process which generates it (see Section \ref{sec:setting-up}).

\subsection{Notation}
We define a neural network image classification model $f(\vec{x};D_\mathrm{t})$, which maps an image $\vec{x}$ to a vector $\vec{y}$ where each element represents the conditional probability of the class label $y$ given the image $\vec{x}$, $D_\mathrm{t}=\{\vec{x}^i, y^i\}^n_{i=0}$ is the training data where $\vec{x}_p \in \mathbb{R}^{n \times n}$, and $y$ is the ground truth class label of $\vec{x}$. $D_\mathrm{t}$ may or may not contain a datum $\vec{x}_p$ with a unique feature $\vec{z}_\mathrm{u} \in \mathbb{R}^{m \times m}$ with $m < n$. We define an additional random feature $\vec{z}_\mathrm{r} \sim U$, with the same dimensionality as $\vec{z}_\mathrm{u}$. We define three datasets $D_\mathrm{c}=\{\vec{x_c}^i\}^n_{i=0}$, $D_\mathrm{u}=\{\vec{x_u}^i\}^n_{i=0}$, $D_\mathrm{r}=\{\vec{x_r}^i\}^n_{i=0}$ out-of-distribution to $D_\mathrm{t}$ which we use to perform inference on $f$ to learn about $\zu$. We make use of the KL divergence between two discrete probability distributions to measure the difference in the model's outputs, $D_{\mathrm{KL}}(P\vert\vert Q)= \mathbb{E}_{\mathrm{x}\sim P}\big[\log{\frac{P(x)}{Q(x)}}\big]$.

\subsection{Method to measure unique feature memorisation}\label{sec:m-score}

The central idea to approximate the memorisation of $\zu$ is to perform a set of inferences on $f$ using out-of-distribution image pairs which are clean,\footnote{Clean means that the image does not contain $\zu$ or $\zr$.} or contain $\vec{z}_r$ or $\vec{z}_{u}$.

\begin{figure}[t]
\centering     
\subfigure[$\xc$]{\label{fig:a}\includegraphics[width=0.3\linewidth]{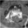}}
\subfigure[$\xu$]{\label{fig:b}\includegraphics[width=0.3\linewidth]{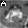}}
\subfigure[$\xr$]{\label{fig:c}\includegraphics[width=0.3\linewidth]{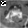}}
\caption{Example image pairs from greyscale CIFAR-10 used for inference on $f$ trained on MNIST or Fashion-MNIST.}
\label{fig:inference-images}
\end{figure}

We define three datasets for inference $D_\mathrm{c}$, $D_\mathrm{u}$, $D_\mathrm{r}$. The datasets we use are out-of-distribution (OOD) to the training dataset. I.e. they are from a different domain to $D_t$. We assume that any dataset which is not the training set, $D_\mathrm{t}$, is OOD and can be used for inference. This is because our method finds only the relative distances between model outputs from image pair inputs and as such the specific distribution which the model outputs is not important. Each dataset is a replica of the base dataset $D_c$ with the exception that every image in $D_u$ also contains $\zu$ and every image in $D_r$ contains $\zr$, where $\zr$ is drawn randomly for every image. Figure \ref{fig:inference-images} shows examples of $\xc$, $\xu$, $\xr$.

We assume that an $f$ which has memorised $\zu$ will, on average, be more sensitive to images that contain $\zu$ than $\zr$. We define memorisation as any \textit{learning} which takes place on $\zu$. Since $\zu$ is unique and cannot be learnt from any other label structure in the training data, any learning which does occur must be memorisation.

We assume that $f$ has been trained with a softmax activation on the final layer under maximum-likelihood such that $f(\vec{x}_\mathrm{c})$, $f(\vec{x}_\mathrm{u})$, $f(\vec{x}_\mathrm{r})$ are valid probability distributions, and such that the KL divergences of any combination are valid. Consider measuring the network outputs of $f$ for every image pair in the OOD inference datasets $D_\mathrm{c}$, $D_\mathrm{u}$, $D_\mathrm{r}$; we calculate the following divergences:
\begin{equation}
    d_u^i = D_{\mathrm{KL}}\big(
    f(\vec{x}_\mathrm{c}^i)
    \vert\vert
    f(\vec{x}_\mathrm{u}^i)
    \big),
\end{equation}
\begin{equation}
    d_r^i = D_{\mathrm{KL}}\big(
    f(\vec{x}_\mathrm{c}^i)
    \vert\vert
    f(\vec{x}_\mathrm{r}^i)
    \big),
\end{equation}
where $d_u^i$, and $d_r^i$ measure the distances in the predictions from $\xc^i$ to $\xu^i$ and $\xr^i$ where $i$ is a sample index in the datasets $D_u$, $D_c$ and $D_r$.  

When the divergences $d_u^i$, and $d_r^i$ are zero there is no difference between the prediction on the clean image and the random or unique feature image, whereas a network that is more sensitive to $\zu$ than any given $\zr$ will have $d_u^i > d_r^i$.


Since $\zr$ is random it is possible that $\zr$ is already a feature learnt by the classifier, and thus $d_u^i < d_r^i$. However, we assume that the subspace of such features is far smaller than the randomness space of $\zr$, and thus most examples that we draw will not be features learnt by the classifier. Also for some $\zr$ the network may be more or less sensitive due to the lack of robustness of CNNs to out-of-distribution data \citep{shao2020understanding}. Therefore, to mitigate the similarity of random features to those learnt by the classifier and robustness issues, we measure the sensitivity of the network by taking the average over every $\zr$ and $\xc$. We abuse the notation and assume that each draw of $\zr$ results in an $\xr$ with a new random unique feature.
\begin{equation}
    X_\mathrm{u} = \mathbb{E}_{\mathrm{\vec{x}_c}\sim P}
    \Big[
    D_{\mathrm{KL}}\big(
    f(\vec{x}_\mathrm{c})
    \vert\vert
    f(\vec{x}_\mathrm{u})
    \big)
    \Big],
    \label{eq:Xu}
\end{equation}
\begin{equation}
    X_\mathrm{r} = 
    \mathbb{E}_{\mathrm{\vec{z}_r}\sim U}
    \Big[
    \mathbb{E}_{\mathrm{\vec{x}_c}\sim P}
    \Big[
    D_{\mathrm{KL}}\big(
    f(\vec{x}_\mathrm{c})
    \vert\vert
    f(\vec{x}_\mathrm{r})
    \big)
    \Big]
    \Big],
    \label{eq:Xr}
\end{equation}
%
where $P$ is the data distribution of the base OOD dataset. We assume that drawing $\xc$ also draws the image pair $\xu$ with the unique feature. We approximate equations \ref{eq:Xu} and \ref{eq:Xr} by sampling from a benchmark dataset that is not the training dataset, and by $\vec{z}_r\sim U$ for every image in the OOD dataset. For memorisation tests on MNIST and Fashion-MNIST we use a greyscale version of CIFAR-10 for inference, and for tests on CIFAR-10 we use a 3-channel version of the MNIST dataset. 

\noindent \textbf{The score $M$} We define our memorisation score for $\zu$ as $M = X_u - X_r$, where $M>0$ corresponds to memorisation of the unique feature. To illustrate the statistical significance of the differences in the means we use a one-tail t-test with the alternative hypothesis that $X_u > X_r$.

\subsection{Setting up canaries}

\subsubsection{Measuring the memorisation of canaries}\label{sec:setting-up}
We use the following test setup to approximate the memorisation of unique features. An ideal test of the memorisation score is to measure the memorisation of a feature on an image that we know contains a unique feature. Since we do not know a priori which unique features have been accidentally left in the training data by a fallible pre-processor, we adopt the strategy of placing canaries into the training data artificially \citep{carlini2019secret}. Once a training example is identified as a canary, the model is trained on the augmented dataset. Next, the memorisation of the unique feature is approximated using the memorisation score method.

\begin{figure}[t]
\centering     
\subfigure[MNIST canary with unique feature \textit{A}]{\label{fig:a2}\includegraphics[width=0.28\linewidth]{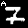}}\hfill
\subfigure[Fashion-MNIST canary with unique feature \textit{A}]{\label{fig:b2}\includegraphics[width=0.28\linewidth]{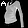}}\hfill
\subfigure[CIFAR-10 canary with unique feature \textit{A}]{\label{fig:c2}\includegraphics[width=0.28\linewidth]{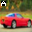}}
\caption{Canaries for testing unique feature memorisation. Each canary has an \textit{A} burnt into its top-left corner.}
\label{fig:canaries}
\end{figure}

The experiments in this work use as \textit{unique feature} a tiny patch of the letter `\textit{A}'. This patch is $5\times5$ pixels, and it is inserted into the top-left corner one pixel from the top-left corner of the canary image. Examples of the unique feature, \textit{A}, embedded in canary images are shown in Figure \ref{fig:canaries}. We assume that the service knows the typography of the letter \textit{A}, its size and its location in the image. For some applications, this is a reasonable assumption since these properties of the feature are public. For example, consider a personal ID verification classifier. Although the name and address of an ID owner are private, the typography, size and location of the private information are the same on every ID and therefore public.

\subsubsection{Selecting high self-influence examples as canaries}

We do not expect that the unique feature will be memorised for every canary since not all training examples are memorised by neural networks \citep{Krueger2019}. Therefore it is unknown a priori which canaries will have unique features that will be memorised. This makes it challenging find to find true positives to test our score against.

We do not want to search every example for the possibility of unique feature memorisation for computational reasons. Additionally, we cannot add unique features to every training example and train once since this will alter the characteristics of the dataset. 
Therefore we should choose efficiently which examples should have the unique feature. Here we develop a strategy to choose examples based on a self-influence score for each example.

It has been shown that memorisation of training labels occurs when training examples from sub-populations of a class are mislabelled, or are members of an under-sampled class \citep{feldman-what-neural-networks}. It is also expected that sub-population or mislabelled examples are highly-influential. This is because they have unique features which require memorisation to classify them correctly \citep{Harutyunyan2021}.

We suppose that unique features are more likely to be memorised when they are present on examples that are highly influential to predictions made on themselves. Conversely, we expect that if $\zu$ is present on a low-influence sample it is unlikely to be memorised since the sample can be predicted by features in similar low-influence samples.  Furthermore the information bottleneck (IB) principle suggests that the network should learn only relevant, and minimal, information (representations) between the input and the output task. Thus, according to the IB principle, a network should ignore unique, features which appear random, \citep{7133169, Achille2018}, as such features will be considered nuisance.  
Finally, it is possible that a unique feature may contribute to shortcut learning \citep{Geirhos2020-shortcut-learning} as we touch on in our discussion.


To determine high-influence examples we use TracIn \citep{Garima2020}. TracIn measures the self-influence of an example by computing the training loss on itself between successive iterations of stochastic gradient descent (SGD) where the loss is computed for that example. Large successive differences in loss will be associated with samples whose predictions can only be improved upon by the network learning from that example or a small subset of similar examples. Small changes will be associated with low-influence examples since learning will be spread over many examples with correlated features. TracIn is approximated for mini-batch (SGD) by measuring the sums of the squared gradients of the loss evaluated on the test example over several checkpoints (see equation \ref{tracincp}). We use TracIn as opposed to other influence functions as it is extremely computationally efficient. The function for computing TracIn using checkpoints is given by

\begin{equation}
    \textrm{TracInCP} = \sum {\eta_i} ||\nabla l (\vec{w}_i, \vec{x})||^2,
    \label{tracincp}
\end{equation}
where $\vec{w}_i$ and $\eta_i$ are the weights and learning rate of $f$ and the optimisation algorithm at checkpoint $i$. $\vec{x}$ is the self-influence test image.

For each dataset/model combination in this work, we measure the self-influence of all training examples by selecting 10 evenly spaced checkpoints which account for a 95\% reduction in the training loss \citep{Garima-2022-faq}. We select the top-15 and bottom-15 examples as canaries. We create 30 models, each of which is trained on the canary with the addition of a unique feature.

\subsubsection{Early Stopping}
It is well known that over-parameterised neural networks memorise random training labels when they are trained indefinitely \citep{Zhang2021-understanding-mem}. However, it has also been shown that LLMs memorise unique phrases even before overfitting in an average sense occurs \citep{carlini2019secret}. We refer to models trained without overfitting on average as well-trained. We suggest that unique features are also memorised by image classification models even when these models are well-trained. Specifically, we test this hypothesis by making use of early stopping during model training with unique features. All models are trained for 500 epochs with 10 patient epochs, and after training, we select the model checkpoint which has the lowest validation loss. We train models using the Adam optimiser and with a cross-entropy loss function \citep{kingma2017adam}.

\section{Results} \label{section:results}

Our primary aim is to show that MLPs and CNNs memorise unique features.
In Section \ref{result:1} we start showing that the memorisation score is capable of detecting lack of memorisation and actual memorisation in experiments with MLPs
on MNIST \citep{lecun2012mnist} and Fashion-MNIST \citep{xiao2017fashion}. We then proceed (Section \ref{result:2}) to show that unique features are also memorised in MNIST, Fashion-MNIST and CIFAR-10 \citep{krizhevsky2009-cifar10} for three common CNN architectural styles. One might expect that explicit and implicit regularisers which prevent overfitting should also reduce memorisation; however in Section \ref{result:3} we show results to the contrary. Finally, in Section \ref{result:4} we discuss the correlation between unique feature memorisation and the self-influence of training examples.

\subsection{Memorisation of unique features in MLPs}\label{result:1}
\subsubsection{Baseline memorisation test on MNIST}

We show in an experiment on MNIST that the $M$ score is less than zero for all canaries when the unique features are omitted from the training data, and that it can be greater than zero for canaries when unique features are memorised.

We use an MLP with two dense layers with 512, 256 outputs respectively, and non-linear ReLU activation functions on each layer. We train with a learning rate of $\num{1e-4}$ and a batch size of 128.

We consider two setups: the first where the training dataset is clean, and the second where we introduce a canary on a random image sampled from the training data. We repeat each setup 15 times.


We find no cases where $M>0$ on the clean dataset. This expected result suggests that our memorisation score has high specificity. When we introduce features we find eight canaries (with injected unique features) with $M>0$. (These results are statistically significant using the method in Section \ref{sec:m-score}.) We do not expect that a unique feature will be memorised for every canary; hence, we do not include canaries (with injected features) that have $M\le 0$.


\subsubsection{MLP memorisation test results}

Next, we evaluate unique feature memorisation in MNIST and Fashion-MNIST for MLP models. The setup of the network is the same as before.

Table \ref{table:mlp-memorisation} shows the memorisation scores for canaries. Each row shows the result for a \textit{new} model trained on a dataset containing a single canary. First, we show the scores for the top-5 most memorised canaries over the several models we trained, and second, we show the average value of the memorisation score over canaries having $M>0$ belonging to a particular dataset. Such averaging isolates instance-based variation and allows us to investigate how the memorisation of memorised unique features varies across approaches.

Top-5 results clearly show that memorisation occurs for unique features in the canaries since $M>0$. We emphasise here that for each result memorisation occurs when \textit{only} a single unique feature has been included in the training data, and that there are no unique features in the test data.


We will use the average memorisation score as a comparison between model architectures where the choice of canaries varies. The variation occurs because canaries are chosen by the self-influence score which depends on the model architecture.

\begin{table}[t]
\centering
\caption{Memorisation scores ($M$) of unique features in MLPs trained on MNIST and F-MNIST. All results in bold correspond to test statistics with p-values $<0.05$. The bottom two rows indicate values of $M$ which have been averaged over canaries with $M>0$.}
\label{table:mlp-memorisation}
\begin{tabular}{lcrrr}
\toprule
IMAGE ID & DATASET &  $X_u$ &  $X_r$ &             $M$ \\
\midrule
   27225 &   MNIST &  0.021 &  0.017 & \textbf{0.0038} \\
    6885 &   MNIST &  0.027 &  0.025 & \textbf{0.0021} \\
   27155 &   MNIST &  0.011 & 0.0094 & \textbf{0.0017} \\
   11708 &   MNIST &   0.01 & 0.0089 & \textbf{0.0012} \\
    8898 &   MNIST & 0.0088 & 0.0081 & \textbf{0.0007} \\
   37251 & F-MNIST &  0.058 &  0.028 &   \textbf{0.03} \\
    2731 & F-MNIST &  0.047 &  0.026 &  \textbf{0.021} \\
    2181 & F-MNIST &  0.039 &  0.021 &  \textbf{0.019} \\
    3694 & F-MNIST &  0.051 &  0.033 &  \textbf{0.019} \\
   16002 & F-MNIST &  0.059 &  0.046 &  \textbf{0.013} \\
   \midrule
    AVERAGE &   MNIST &  0.013 &  0.012 & {0.0013} \\
    AVERAGE & F-MNIST &  0.033 &  0.024 & {0.0093} \\
\bottomrule
\end{tabular}
\end{table}

\subsection{Memorisation of unique features in CNNs}\label{result:2}

We extend our investigation to CNN architectures on the same datasets. We test for memorisation on three architectures: \textit{CNN-1} a simple two convolutional layer CNN; \textit{CNN-2} a small VGG-type CNN \citep{simonyan2015-vgg}; and \textit{DenseNet} a DenseNet with 100 layers \citep{Huang2017-densenet}.\footnote{Only four canaries were chosen in DenseNet due to computational constraints.} See Appendix \ref{app:net-arch} for details on the architectures and their training. 

Table \ref{table:cnn-memorisation} shows memorisation scores for canaries. We show first the scores for the top-5 most memorised canaries over the several models we trained, and second the average value of the memorisation score for all memorised canaries belonging to a particular dataset.

It can be seen from the top-5 data in Table \ref{table:cnn-memorisation} that memorisation occurs in all network architectures for each of the datasets tested (since $M>0$). 

We now compare with the results in Table \ref{table:mlp-memorisation} for the MLP architecture. 

The average memorisation is greater in the CNN architecture than the MLP architecture for MNIST ($0.012 > 0.0013$) and Fashion-MNIST ($0.028 > 0.0093$). In fact, the average memorisation in the CNNs is over an order of magnitude greater for MNIST. We also cautiously suggest that the average memorisation is greater for CIFAR-10 in DenseNet than in CNN-2 ($0.38 > 0.018$) though the sample size for the DenseNet results is small.

\citet{feldman-what-neural-networks} suggests that example memorisation is closely linked to model accuracy. We suggest that the increases in memorisation seen in the CNNs and the DenseNet can also be explained by the model's accuracy. For example, DenseNet has a validation accuracy of~92\% and CNN-2~72\%.

\begin{table}[t]
\centering
\caption{Memorisation ($M$) scores of unique features in CNN-1, and CNN-2, a DenseNet trained on MNIST, Fashion-MNIST and CIFAR-10. Bold denotes statistical significance.}
\label{table:cnn-memorisation}
\begin{tabular}{lccr}
\toprule
   ID &  DATASET &    MODEL &            $M$ \\
\midrule
51508 &    MNIST &    CNN-1 & \textbf{0.073} \\
14873 &    MNIST &    CNN-1 & \textbf{0.035} \\
 7080 &    MNIST &    CNN-1 & \textbf{0.021} \\
43454 &    MNIST &    CNN-1 &  \textbf{0.02} \\
47034 &    MNIST &    CNN-1 &  \textbf{0.02} \\
59677 &  F-MNIST &    CNN-1 & \textbf{0.085} \\
23711 &  F-MNIST &    CNN-1 & \textbf{0.068} \\
15748 &  F-MNIST &    CNN-1 & \textbf{0.059} \\
12168 &  F-MNIST &    CNN-1 & \textbf{0.058} \\
10477 &  F-MNIST &    CNN-1 &  \textbf{0.05} \\
23308 & CIFAR-10 &    CNN-2 & \textbf{0.058} \\
 9461 & CIFAR-10 &    CNN-2 & \textbf{0.023} \\
 7371 & CIFAR-10 &    CNN-2 & \textbf{0.018} \\
35174 & CIFAR-10 &    CNN-2 & \textbf{0.017} \\
15726 & CIFAR-10 &    CNN-2 & \textbf{0.013} \\
32574 & CIFAR-10 & DenseNet &  \textbf{0.72} \\
  772 & CIFAR-10 & DenseNet &  \textbf{0.25} \\
 8022 & CIFAR-10 & DenseNet &  \textbf{0.18} \\
 \midrule
 AVERAGE &    CNN-1 &    MNIST & {0.012} \\
 AVERAGE &    CNN-1 &  F-MNIST & {0.028} \\
 AVERAGE &    CNN-2 & CIFAR-10 & {0.018} \\
 AVERAGE & DenseNet & CIFAR-10 &           0.38 \\
\bottomrule
\end{tabular}
\end{table}


\subsection{Effects of explicit and implicit regularisation on memorisation}\label{result:3}

Typically regularisation is used to reduce a model's ability to overfit training data. It has been shown that explicit and implicit regularisation strategies do not reduce the ability of a network to fit random training data \citep{Arplt2017-a-closer-look-at-memorization}. This suggests that they should also have a low influence on the memorisation of unique features. This experiment explores exactly this: i.e., whether these strategies eliminate the risk of memorisation of unique features in networks that on average are not overfitted. We use three regularisation strategies: dropout, data augmentation, and batch normalisation, applied in various combinations. After training, we measure the memorisation of each of the canaries.

Table \ref{table:regularisation} shows the average memorisation score overall canaries where $M>0$ for MNIST, Fashion-MNIST and CIFAR-10 trained by two network architectures. 
Whilst the top-5 scores have been omitted from the table they show that the regularisation methods do not eliminate memorisation of unique features since $M > 0$ for every canary. The mean scores in this table can be compared with the MLP and CNN data in Tables \ref{table:mlp-memorisation} and \ref{table:cnn-memorisation}.



The results, taking MNIST for example, illustrate that almost all values are higher than the average of 0.0013 obtained from Table \ref{table:mlp-memorisation} without any regularisation. Canaries in CIFAR-10 have greater memorisation scores than in the non-regularised models shown in Table \ref{table:cnn-memorisation}. These results extend to the level of features, findings that have been made previously for the memorisation of whole training examples \citep{Zhang2021-understanding-mem}. 

Data augmentation is a particularly interesting case. When CNN-2 is trained on CIFAR-10 with data augmentation the average memorisation increases from 0.018 to 0.13. This is interesting because many datasets require data augmentation to achieve the best accuracy. We offer below a potential explanation as to why such an increase happens. 
%

Data augmentation is employed to improve the learning of translational and/or rotational in/equi-variances to image features. Perturbation of canaries implicitly adds additional canaries to the dataset, and inadvertently increase the spurious correlation between the unique feature and its label. This enables easier learning of the unique feature since the collective contribution to the training loss from the canaries is greater during training.


\begin{table}[t]
\centering
\caption{Average memorization ($M$) scores over canaries with $M>0$ for unique features for models with explicit and implicit regularisers, such Dropout, Data Augmentation (Augm.), and Batch Normalization (Batch Norm.).}
\label{table:regularisation}
\begin{small}
\begin{tabular}{lclr}
\toprule
DATASET & MODEL &       REGULARISATION &             $M$ \\
\midrule
       MNIST &   MLP &       Dropout &            0.02 \\
       MNIST &   MLP &       Augm. &          0.0033 \\
       MNIST &   MLP & Dropout \& Augm. &          0.0013 \\
       MNIST &   MLP &       Batch Norm. &           0.039 \\
     F-MNIST &   MLP &       Dropout &           0.025 \\
     F-MNIST &   MLP &       Augm. &          0.0065 \\
     F-MNIST &   MLP & Dropout \& Augm. &           0.029 \\
     F-MNIST &   MLP &       Batch Norm. &         0.00097 \\
    CIFAR-10 &   CNN-2 &       Dropout &            0.38 \\
    CIFAR-10 & CNN-2 &       Augm. &            0.13 \\
    CIFAR-10 & CNN-2 & Dropout \& Augm. &            0.44 \\
    CIFAR-10 & CNN-2 &        Batch Norm. &             1.7 \\
\bottomrule
\end{tabular}
\end{small}
\end{table}


\subsection{Unique feature memorisation for high/low self-influence examples}\label{result:4}

In the previous experiments, we selected canaries based on self-influence. To examine the relationship between memorisation and self-influence further we measure Pearson's correlation coefficient.

\begin{figure}
    \centering
    \includegraphics{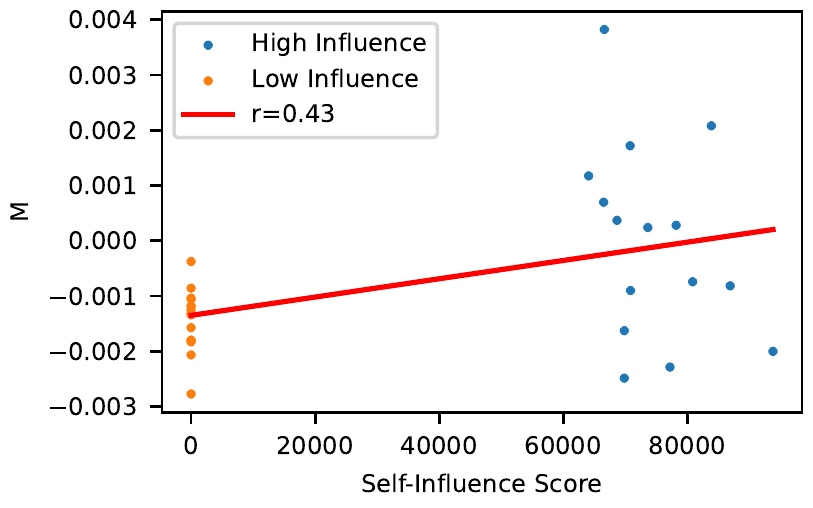}
    \caption{Memorisation scores and self-influence scores for canaries fitted by MLP on MNIST.}
    \label{fig:mnist-correlation}
\end{figure}

Figure \ref{fig:mnist-correlation} shows the memorisation scores and self-influence scores for the top-15 and bottom-15 canaries by self-influence score on MNIST for the MLP. The correlation coefficient is 0.43, indicating a weak correlation between memorisation score and self-influence. However, we did not find a strong correlation for F-MNIST and CIFAR-10 using CNN-1. Instead, the results showed that there was a high degree of unique feature memorisation in the low self-influence examples. We comment further on these results in the discussion section.

\subsection{Discussion}\label{result:disc}


Our main finding is that commonly used neural networks architectures memorise unique features when the unique feature occurs once in training data for several scenarios.

We have shown that in the MNIST dataset a correlation exists between the self-influence score (measured using TracIn) and the memorisation score. Contrary to our expectations, we did not find a significant correlation in our experiments for Fashion-MNIST and CIFAR-10. We suspect that one reason is the influence of short-cut learning. For example, under the \textit{Principle of Least Effort}, the network may find it easier to fit the unique feature, than to learn features for training examples from Fashion-MNIST or CIFAR-10 \citep{Geirhos2020-shortcut-learning}. For MNIST the task of learning is significantly easier and the unique feature is only memorised on very high influence examples. Another possibility is that TracIn does not trace examples that are necessarily memorable in datasets with more complex features, and hence the unique features are not memorised either.

\section{Conclusion}

In this work, we present a memorisation score to measure the memorisation of unique features in imaging datasets by neural network image classification models. We focus on the case where the unique feature appears on a single image in the training data, and where we have access only to the model's inputs and outputs, the unique feature, and no access to the training data except for its domain.

We show that unique feature memorisation does occur in this setting, and is not eliminated by typical explicit and implicit regularisation strategies, dropout, data augmentation and batch normalisation. We derive these results for benchmark datasets and a range of neural network architectures.

The results, even in standard benchmark datasets, suggest that neural networks pose a privacy risk to unique sensitive information in imaging datasets even if the information occurs once. The information does not have to be rare in the wild. In the context of a healthcare application, the information could be a patient name that was not removed by an image pre-processor. 


\section{Acknowledgements}
This work is supported by iCAIRD, which is funded by Innovate UK on behalf of UK Research and Innovation (UKRI) [project number 104690]. S.A.\ Tsaftaris acknowledges also support by a Canon Medical / Royal Academy of Engineering Research Chair under Grant RCSRF1819\textbackslash8\textbackslash25. This work
was partially supported by the Alan Turing Institute under EPSRC grant EP/N510129/1.


\bibliography{main.bib}
\bibliographystyle{apalike}

\newpage
\appendix
\onecolumn

\section{Network Architectures}\label{app:net-arch}

We evaluate our memorisation score using several common architectural styles of neural networks. The first, \textit{CNN-1} is trained on MNIST and Fashion-MNIST datasets. It is comprised of: \textrm{Conv2D(32,3,3)} $\rightarrow$ \textrm{ReLU} $\rightarrow$ \textrm{Conv2D(64,3,3)} $\rightarrow$ \textrm{MaxPool2d(2,2)} $\rightarrow$ \textrm{ReLU} $\rightarrow$ \textrm{Dense(128)} $\rightarrow$ \textrm{ReLU} $\rightarrow$ \textrm{Dense(128)} $\rightarrow$ \textrm{ReLU} $\rightarrow$ \textrm{Dense(\#classes)} $\rightarrow$ \textrm{Softmax}.

We train CNN-1 with a learning rate of $\num{3e-4}$ and a batch size of 128.

\textit{CNN-2} is small VGG-style network trained on CIFAR-10. It is comprised of: \textrm{Conv2D(32,3,3)} $\rightarrow$ \textrm{ReLU} $\rightarrow$ \textrm{Conv2D(32,3,3)} $\rightarrow$ \textrm{ReLU} $\rightarrow$ \textrm{MaxPool2d(2,2)} $\rightarrow$ \textrm{Conv2D(64,3,3)} $\rightarrow$ \textrm{ReLU} $\rightarrow$ \textrm{Conv2D(64,3,3)} $\rightarrow$ \textrm{ReLU} $\rightarrow$ \textrm{MaxPool2d(2,2)} $\rightarrow$ \textrm{Dense(1024)} $\rightarrow$ \textrm{ReLU} $\rightarrow$ \textrm{Dense(\#classes)}.

We train CNN-2 with a learning rate of \num{3e-4} and a batch size of 512.

\textit{DenseNet} is DenseNet trained on CIFAR-10 \cite{huang2018densely}. The network has 100 layers, a growth factor of 12, and three dense blocks. We use an existing implementation and train using the same parameters given in \citet{Atienza}.

\section{Experimental codes}
The code for the experiments is forthcoming \texttt{https://github.com/jasminium/unintended-memorisation}.

\end{document}